\documentclass[letterpaper, 10 pt, conference]{ieeeconf}
\IEEEoverridecommandlockouts
\usepackage{amsfonts}       
\usepackage{booktabs}       
\usepackage[T1]{fontenc}    
\usepackage[utf8]{inputenc} 
\usepackage{nicefrac}       
\usepackage{microtype}      
\usepackage{url}            

\usepackage{amsmath}
\usepackage{amssymb}
\usepackage{epsfig}
\usepackage{graphics}
\usepackage{graphicx}
\usepackage{lineno}
\usepackage{tikz}
\usepackage{wrapfig}
\usepackage{xcolor}

\usepackage{blkarray}
\usepackage{makecell}

\usepackage[english]{babel}
\usepackage{algorithm}
\usepackage[noend]{algpseudocode}

\title{\LARGE \bf Anomaly Detection With Conditional Variational Autoencoders}
\author{Adrian Alan Pol$^{1,2}$,
        Victor Berger$^2$,
        Gianluca Cerminara$^1$,
        Cecile Germain$^2$,
        Maurizio Pierini$^1$\\\\%
        \parbox{3 in}{\centering \small
            $^{1}$ European Organization for Nuclear Research (CERN)\\
            Meyrin, Switzerland
        }
        \hspace*{0.5 in}
        \parbox{3 in}{\centering \small
            $^{2}$ Laboratoire de Recherche en Informatique (LRI)\\
            Universit\'{e} Paris-Saclay, Orsay, France
        }
}

\begin{document}

\maketitle

\thispagestyle{empty}
\pagestyle{empty}

\begin{abstract}
Exploiting the rapid advances in probabilistic inference, in particular variational Bayes and variational autoencoders (VAEs), for anomaly detection (AD) tasks remains an open research question. Previous works argued that training VAE models only with inliers is insufficient and the framework should be significantly modified in order to discriminate the anomalous instances. In this work, we exploit the deep conditional variational autoencoder (CVAE) and we define an original loss function together with a metric that targets hierarchically structured data AD. Our motivating application is a real world problem: monitoring the trigger system which is a basic component of many particle physics experiments at the CERN Large Hadron Collider (LHC). In the experiments we show the superior performance of this method for classical machine learning (ML) benchmarks and for our application.
\end{abstract}

\section{Introduction}

AD is expected to evolve significantly due to two factors: the explosion of interest in representation learning and the rapid advances in inference and learning algorithms for deep generative models. Both go well beyond the traditional fully supervised setting, which is generally not applicable for most AD tasks.  Particularly relevant is the variational learning framework of deep directed graphical model with Gaussian latent variables i.e. variational autoencoder (VAE), and deep latent Gaussian model, introduced by \cite{kingma2013auto, pmlr-v32-rezende14}. 

Relatively little work has been devoted to exploit for AD the advances in modeling complex structured representations that perform probabilistic inference effectively. In most of them (discussed in section~\ref{sec:relw}), it has been argued that vanilla VAE architectures may not be adequate for AD, and that they must be specifically tweaked for specific sub-cases of AD with complex extensions. 

This work is motivated by a real world problem: improving AD for the trigger system, which is the first stage of event selection process in many experiments at the LHC at CERN. To be acceptable in this high-end production context, any method must abide to stringent constraints: certainly performance, but also simplicity and robustness, for long-term maintainability.  Because of the nature of our target application the algorithm has to be conditional. In layman terms, some of the structure of the model is known and associated observables are available. This points towards CVAE architectures~\cite{sohn2015learning}. CVAE is a conditional directed graphical model where input observations modulate the prior on latent variables that generate the outputs, in order to model the distribution of high-dimensional output space as a generative model conditioned on the input observation.

The goal of this paper is to explore the relevance of the alleged limitations for an ordinary CVAE, both in a generic setting, and for our specific application. We address two categories of limitations: general issues of (C)VAEs and specific to AD. Our main contributions are as follows.
\begin{itemize}
    \item We  define a new loss function that allows the model to learn the optimal reconstruction resolution.
    \item We design a new anomaly metric associated with the CVAE architecture that provides superior performance on both classical ML and particle physics specific datasets.  
    \item We propose an alternative experimental setup for AD on MNIST dataset.
\end{itemize}{}

The remainder of this paper is organized as follows. Section~\ref{section-problem-statement} outlines the problem we want to solve. Section~\ref{section-background} summarizes the theoretical background, proposed method and related work. We consider a toy experiment, the MNIST and Fashion-MNIST dataset in Section~\ref{section-experiments}. Finally we apply the proposed method to a real problem, related to the monitoring of the CMS experiment at the CERN LHC in Section~\ref{section-experiment-cms}.

\section{Problem Statement}\label{section-problem-statement}

\begin{figure}
    \begin{center}
    \includegraphics[width=0.66\linewidth]{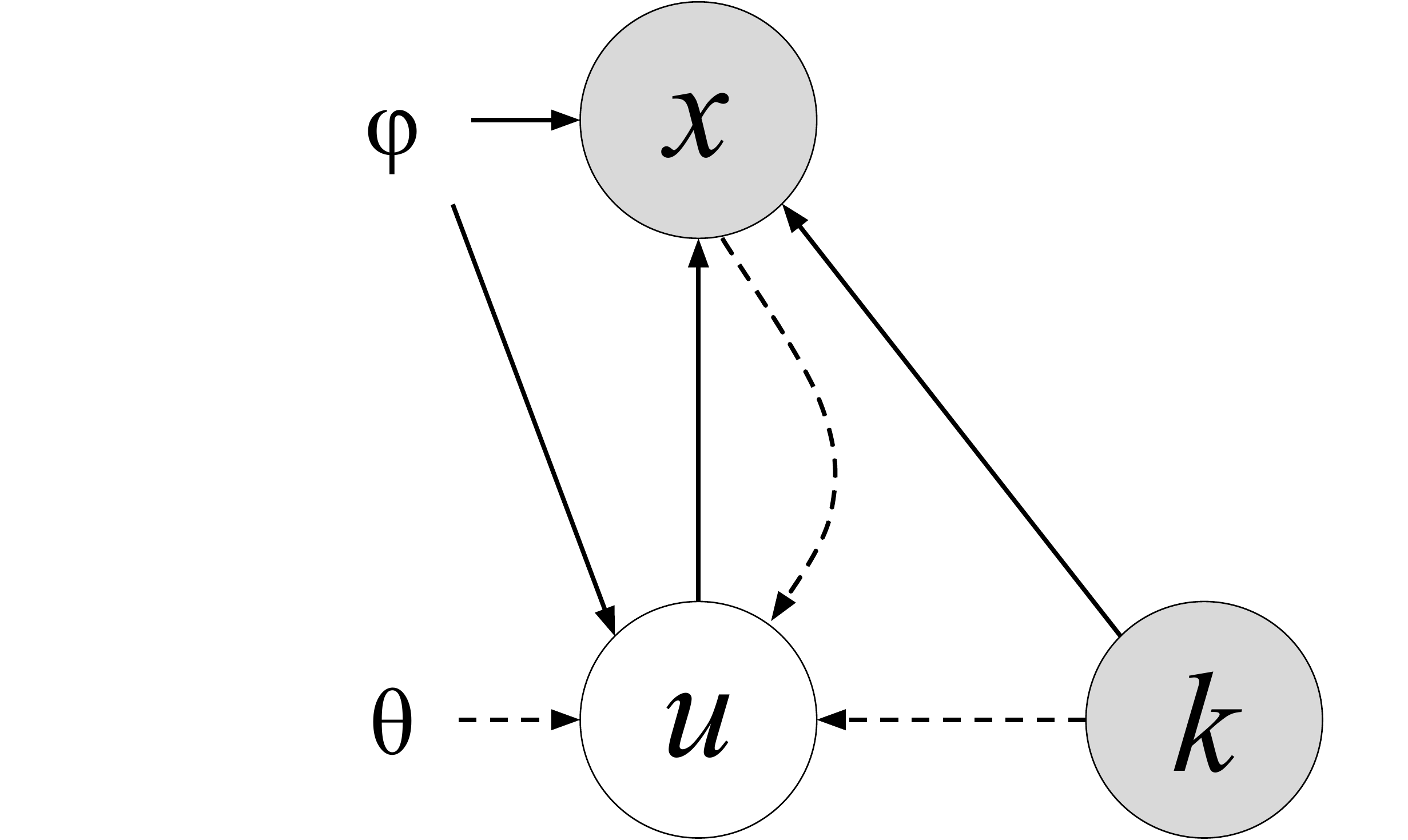}
    \caption{Illustration of CVAE as a directed graph. Solid lines denote the generative model $p_\theta(x | u, k)p_\theta(u)$. Dashed lines denote variational approximation $q_\phi(u|x,k)$. Both variational parameters $\theta$ and generative model parameters $\phi$ are learned jointly.}
    \label{figure-graph}
    \end{center}
\end{figure}

We are operating in a semi-supervised setup, where the examples of anomalous instances are not available. However we know the design of the system and we directly observe some factors of variation in data. The observable $x$ is a function of $k$ ({\em known}) and $u$ ({\em unknown}) latent vectors: $x = f(k, u)$. For a collection of samples $x = [x_{1}, x_{2}, ..., x_{n}]$ we are interested in highlighting instances where we observe:
\begin{itemize}
    \item big change on single feature, we later call {\bf Type A} anomaly, {\it and}
    \item small but systematic change on a group of features in the same configuration group (generated using the same $k$, as we further explain in Section~\ref{section-synthetic}), called {\bf Type B} anomaly.
\end{itemize}
On the contrary, samples with a problem of small severity and on a group of uncorrelated features should be considered as an inlier, purely caused by expected statistical fluctuations.

In summary, we need an algorithm that exploits the known causal structure in data, spots both types of problems listed above, generalizes to unseen cases and uses data instead of relying on feature engineering. Inference time is negligible in the context of the target application (see Section~\ref{section-experiment-cms}).

\section{Background and Proposed Method}
\label{section-background}

\subsection{Variational Autoencoders}
VAEs (\cite{kingma2013auto,pmlr-v32-rezende14}) are a class of likelihood-based directed graphical generative models, maximizing the likelihood of the training data according to the generative model $p_\theta(Data) = \prod_{x \in Data}p_\theta(x)$. To achieve this in a computable way, the generative distribution is augmented by the introduction of a latent variable $z$: $p_\theta(x) = \int p_\theta(x | z)p(z)dz$. This allows to choose $p_\theta(x|z)$ as a simple distribution (like a normal law) and still have the marginal $p_\theta(x)$ to be very expressive, as an infinite mixture controlled by $z$.

The parameter estimation of the graph is problematic due to intractable posterior inference. The VAEs parameters are efficiently trained using an inference distribution $q_\phi(z | x)$ in a fashion very similar to autoencoders, using stochastic gradient variational Bayes framework. The recognition model $q_\phi(z | x)$ is included to approximate the true posterior $p_\theta(z | x))$. Ref. \cite{kingma2013auto} shows that for any such distribution:

\begin{equation}
    \begin{split}
        \log p_\theta(x) - & \mathbb{D}_{\text{KL}}(q_\phi(z | x) \| p_\theta(z | x)) = \\
        &\quad \mathbb{E}_{z \sim q} [\log p_\theta(x | z)] - \mathbb{D}_{\text{KL}}(q_\phi(z | x) \| p(z))
    \end{split}
\end{equation}
Given the Kullback-Leibler (KL) divergence is always positive, the right-hand term of this equality is thus a low-bound of $\log p_\theta(x)$ for all $x$ (called the Evidence Lower Bound, or ELBO). Optimizing it is a proxy for optimizing the log-likelihood of the data, defining the training loss as:

\begin{equation}
     \mathcal{L}_{\text{ELBO}}(x) = \mathbb{E}_{z \sim q} [\log p_\theta(x | z)] - \mathbb{D}_{\text{KL}}(q_\phi(z | x) \| p(z))
\end{equation}

The model choice for $q_\phi(z|x)$, $p(z)$ is generally considered a factorized normal distributions, allowing easy computation of the $\mathbb{D}_{\text{KL}}$ term, and sampling of $z$ through the reparameterization trick~\cite{kingma2013auto}.

It is typical when using VAEs to model the reconstruction as a mean squared error (MSE) between the data $x$ and the output of the decoder. However, this is equivalent to setting the observation model $p_\theta(x|z)$ as a normal distribution of fixed variance $\sigma=1$. Indeed, the log-likelihood of a normal distribution with fixed variance of 1 is given as:

\begin{equation}
    -\log \mathcal{N}(x ; \mu, 1) = \|x - \mu\|^2 + \log(\sqrt{2\pi})
\end{equation}
We argue that fixing the variance this way can be detrimental to learning as it puts a limit on the accessible resolution for the decoder: this defines the generative model as having a fixed noise of variance 1 on its output, making it impossible for it to accurately model patterns with a characteristic amplitude smaller than that. However, unless {\em a priori} knowledge suggests it, there is no guarantee that all patterns of interest would have such a large characteristic amplitude. This is actually trivially false for some very common cases: when the dataset has been normalized to a global variance of 1, or when it is composed of data constrained to a small interval of values, such as images whose pixels are constrained to $[0;1]$. Rather than adding a weighting term between the two parts of the loss like has often been done (\cite{higgins2017beta} for example) we rather let the model learn the variance of the output of the decoder feature-wise ($i$ running as the dimensionality of the data vectors $x$):

\begin{equation}
    - \log p_\theta(x | z) = \sum_i \frac{(x_i - \mu_i)^2}{2 \sigma^2_i} + \log\left(\sqrt{2\pi}\sigma_i\right)
\end{equation}
Learning the variance of the MSE reconstruction allows the model to find the optimal resolution for the reconstruction of each feature of the data, separating the intrinsic noise from the correlations. This empirically gives similar results to associating a fine-tuned weighing parameter, while removing the need to tune said hyper-parameter.

\subsection{Setup Description}

In our setup we have three types of variables, see Figure~\ref{figure-graph}. For random observable variable $x$, $u$ (\textit{unknown}, unobserved) and $k$ (\textit{known}, observed) are independent random latent variables. The conditional likelihood function $p_\theta(x|u,k)$ is formed by a non-linear transformation, with parameters $\theta$. $\phi$ is another non-linear function that approximates inference posterior $q_{\phi}(u | k,x) = N(\mu, \sigma I)$. The latent variables $u$ allow for modeling  multiple modes in conditional distribution of $x$ given $k$ making the model sufficient for modeling one-to-many mapping. To approximate $\phi$ and $\theta$ we optimize the following modified ELBO term:

\begin{equation}
    \begin{split}
        \log p_{\theta}(x) \geq & \mathbb{E}_{q_{\phi}(z | k, x)}[\log p_{\theta}(x | z, k)]\\
        &\quad -\mathbb{D}_{\text{KL}}(q_{\phi}(z|x,k) || p(z))
    \end{split}
\end{equation}
where $z$ (a Gaussian latent variable) intends to capture non-observable factors of variation $u$. The loss is computed as:

\begin{equation}
    \begin{split}
        \mathcal{L}_{\text{CVAE}}(x, k,\theta,\phi) = & \sum_i \frac{(x_i - \mu_i)^2}{2 \sigma^2_i} + \log\left(\sqrt{2\pi}\sigma_i\right) \\ 
        & \quad + \mathbb{D}_{\text{KL}}(q_{\phi}(z|x,k) || p(z)).
    \end{split}
    \label{equation-loss}
\end{equation}

\begin{figure}
    \begin{center}
        \includegraphics[width=\linewidth]{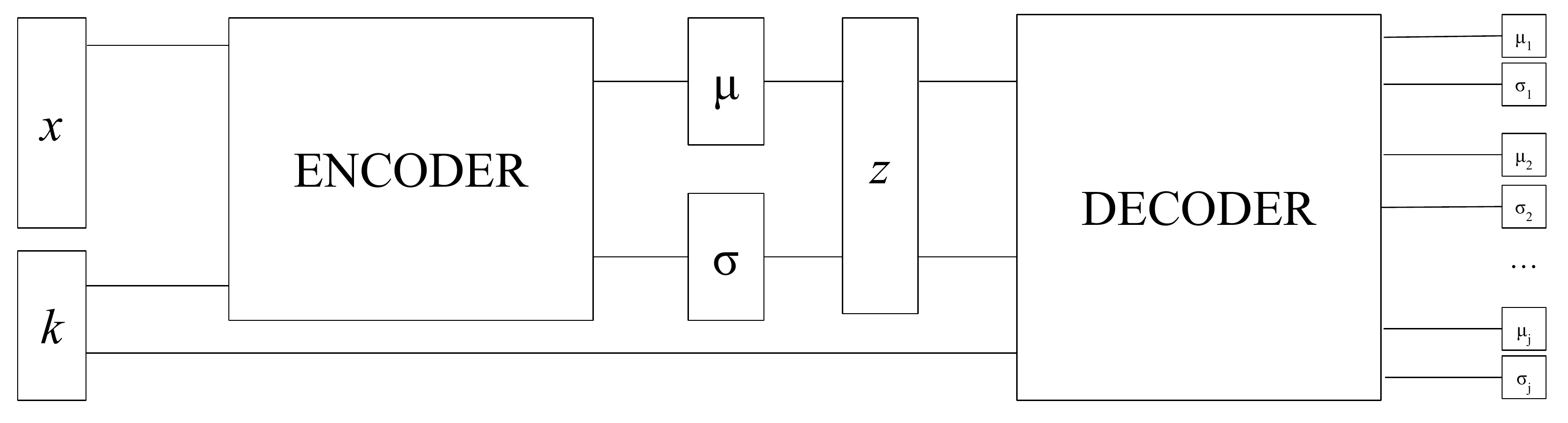}
        \caption{Architecture of CVAE based model for AD.}
        \label{figure-general-architecture}
    \end{center}
\end{figure}

Our model is built upon CVAE framework but we focus on conditional distribution of output variables for AD tasks. We address the difference to VAE setup in Section~\ref{Section-Metric}. The schema of the network architecture, corresponding to a graph from Figure~\ref{figure-graph} is shown in Figure~\ref{figure-general-architecture}. Depending on the experiment, the number and type of hidden layers will vary. We train the model using Keras~\cite{chollet2015keras} with TensorFlow~\cite{abadi2016tensorflow} as a backend using Adam~\cite{kingma2014adam} optimizer and with early stopping~\cite{vincent2010stacked} criterion. Once the model parameters are learned, we can detect anomalies using different metrics:

\begin{itemize}
    \item for Type A with average infinity norm of the reconstruction loss $||\frac{1}{\sigma}(x-\hat{x})^2||_{\infty}$ ($\hat{x}$ as the reconstructed mean and $\sigma$ as the reconstructed variance of decoder output), performing multiple sampling of $z$ (we arbitrarily choose $30$);
    \item for Type B with mean KL-divergence term $\mu(\mathbb{D}_{\text{KL}})$.
\end{itemize}

\subsection{A metric for anomaly detection with CVAE}\label{Section-Metric}
For a given datapoint $(x, k)$, the evaluation of the loss of the VAE at this datapoint $\mathcal{L}(x,k)$ is an upper-bound approximation of $-\log p_\theta(x | k)$, measuring how unlikely the measure $x$ is to the model given $k$. Thresholding the value of this loss is thus a natural approach to AD as explored with good results in \cite{soelch2016variational}. The CVAE thus provides here a model that naturally estimates how anomalous $x$ is given $k$, rather than how anomalous the couple $(x, k)$ is. This means that a rare value for $k$ associated with a proper value for $x$ should be treated as non-anomalous, which is our goal. The CVAE was successfully used for intrusion detection tasks before~\cite{lopez2017conditional}. However authors approach did not use $\mathbb{D}_{\text{KL}}$ as anomaly indicator.

The loss function from Equation~\ref{equation-loss} can be broken up to target two independent problems. Because of two separate failure scenarios we do not combine the metrics in one overall score but rather use logical \texttt{OR} to determine anomalous instances. In the first case we are interested in identifying an anomaly on a single feature. Typically used mean of reconstruction error would likely be an incorrect choice when most of the features do not manifest abnormalities and lower the anomaly score. In the second case we expect $\mu_{z}$ to land on a the tail of the distribution for anomalous cases. As argued in~\cite{gemici2017generative} the $\mathbb{D}_{\text{KL}}$ measures the amount of additional information needed to represent the posterior distribution given the prior over the latent variable being used to explain the current observation. The lower the absolute value of $\mathbb{D}_{\text{KL}}$ the more predictable state is observed.

Finally, the use of the VAE framework guarantees that the method generalizes to unseen observations as argued in~\cite{kingma2014semi}.

\subsection{VAE for anomaly detection}
\label{sec:relw}
Deep architectures have become increasingly popular in semi-supervised AD~\cite{Chalapathy19}. They cope with the issues of the classical methods, $\mu$-SVM~\cite{scholkopf2001estimating}, and Isolation Forest~\cite{liu2012isolation} (IF). As argued by~\cite{bengio2007scaling}, the $\mu$-SVM kernel-based classification does not scale to  high data dimensionality, as it requires that the function to learn be smooth enough to achieve generalization by local interpolation between neighboring examples. Isolation assumes that anomalies can be isolated in the native feature space. 

The need for agnostically learning a representation from the data can be addressed indirectly by deep networks in a classification or regression context~\cite{Tishby:2017}, and be exploited for semi-supervised AD~\cite{Hendrycks17}. Autoencoders are particularly adapted to semi-supervised AD. When trained on the nominal, testing on unseen faulty sample tend to yield sub-optimal representations, indicating that a sample is likely generated by a different process. Until relatively recently, the autoencoding approach was restricted to learning a deterministic map of the inputs to the representation, because the inference step with these representations would
suffer from high computational cost~\cite{bengio2013representation}. A considerable body of work has been devoted to evolve these architectures towards learning density 
models implicitly~\cite{Alain:2014}. 

The dissemination of the generative models, and specifically the VAE, offer a more general and principled avenue to autoencoding-based AD. \cite{AnCho15} describes a straightforward approach for VAE-based AD. It considers a simple VAE, and the Monte-Carlo estimate of the expected reconstruction error (termed reconstruction probability), which is similar to our metric for Type A problem.  Experiments on MNIST and KDD demonstrate a majority of superior performance of VAE over AE and spectral methods. 

However,~\cite{Wang19} argues that the probabilistic generative approach of VAE could suffer from an intrinsic limitations when the goal is AD, with two arguments. Firstly, because the model is trained only on inliers, the representation will not be discriminative, and will essentially overfit the normal distribution. Secondly, the representation might even be useless, falling back to the prior; technically because the generator is too powerful, the regularization by the $\mathbb{D}_{\text{KL}}$ vanishes \cite{Zhao17}. 

\cite{Kawachi18} addresses this issue with specific hypotheses on the distributions of inliers and anomalies. A more general approach~\cite{Hendrycks19,Wang19} is to learn a more conservative representation by exposing the model to out-of-distribution (abnormal) examples, still without knowledge of the actual anomaly distribution, with adversarial architectures and specific regularizations. While~\cite{Hendrycks19} simply defines an ad-hoc regularization and hyperparameter optimization, \cite{Wang19} proposes an adversarial architecture for generating the anomalies and exploiting them to create a less overfitted representation. Neither of these approaches would meet the robustness and simplicity specifications of our motivating application.  

\section{Experiments on benchmarks}\label{section-experiments}

\subsection{Anomaly Detection on MNIST and Fashion-MNIST}

We asses the performance of the proposed method using the MNIST~\cite{lecun1998gradient}, and the more recent and more challenging Fashion-MNIST~\cite{xiao2017fashion} datasets as examples of possible real-world applications. Both datasets contain gray-level images of handwritten digits and pieces of clothing respectively. Handwritten digits in MNIST dataset belong to a manifold of dimension much smaller than the dimension of $x$ ($28$x$28$ pixels), because the majority of random arrangements of pixel intensities do not look like handwritten digits. Intuitively, we would expect this dimension to be at least the size of $10$ as the number of classes suggests. But we need to accommodate larger latent space as each digit can be written in different style. Similar intuition applies to Fashion-MNIST as this dataset also has $10$ target classes of clothing types but there is variability inside a class e.g. type of shoe. 

\begin{figure}
    \begin{center}
        \begin{tabular}{  >{\centering\arraybackslash}m{1.1cm} m{6.4cm} }
            \small CVAE & \includegraphics[width=\linewidth]{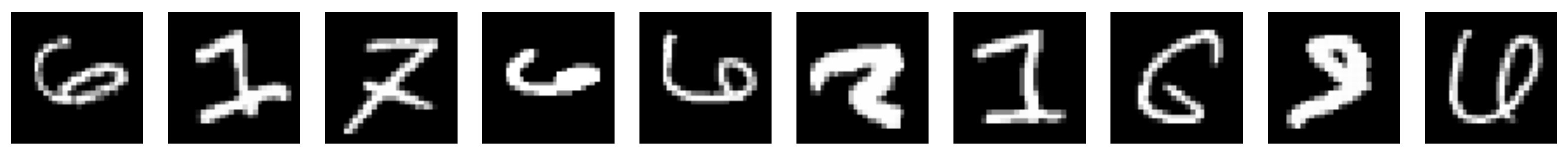}
                          \includegraphics[width=\linewidth]{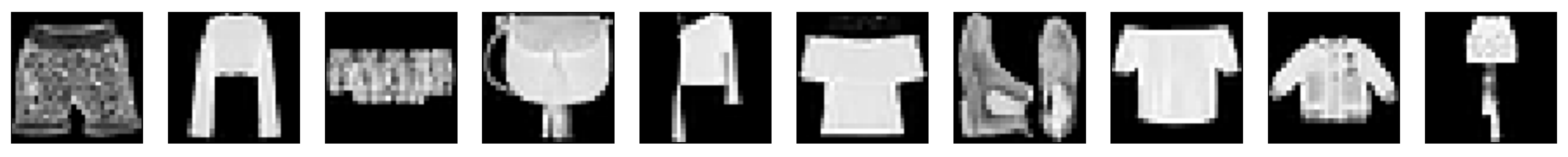} \\
            \small VAE & \includegraphics[width=\linewidth]{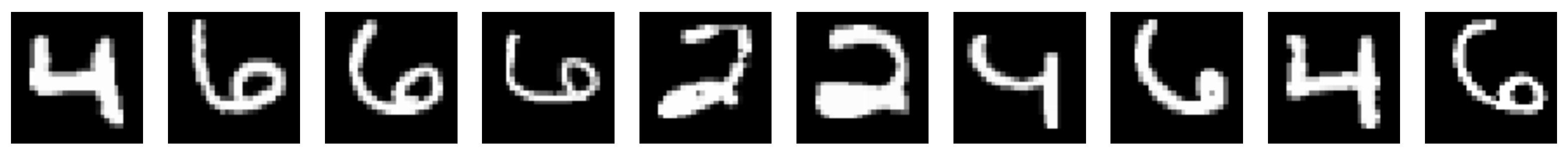}
                         \includegraphics[width=\linewidth]{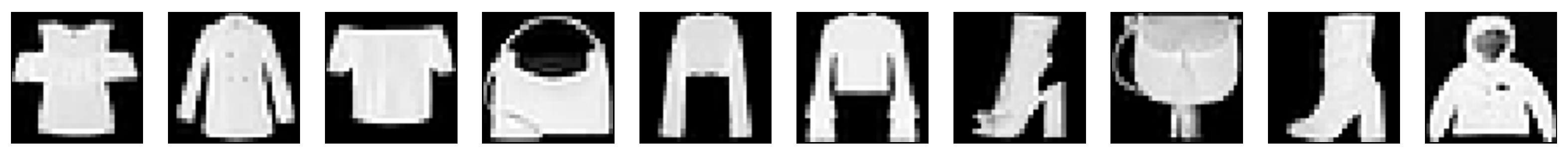} \\
            \small $\mu$-SVM & \includegraphics[width=\linewidth]{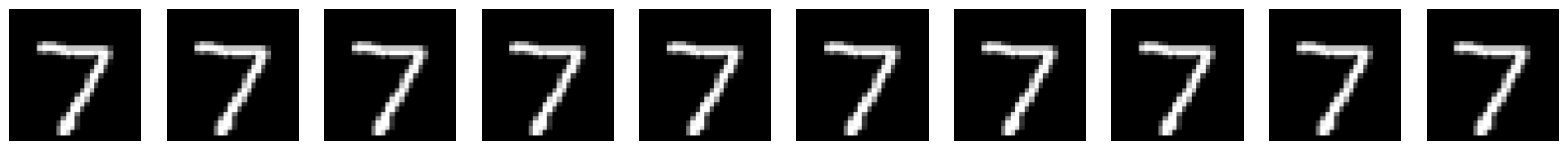}
                               \includegraphics[width=\linewidth]{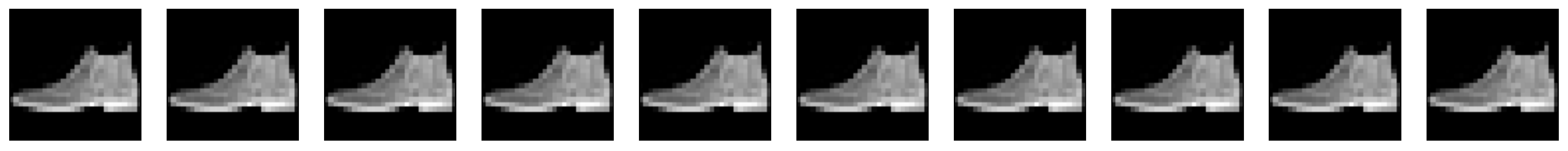} \\
            \small IF & \includegraphics[width=\linewidth]{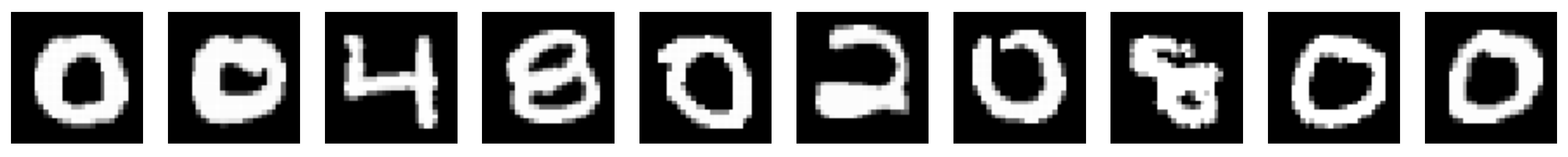}
                        \includegraphics[width=\linewidth]{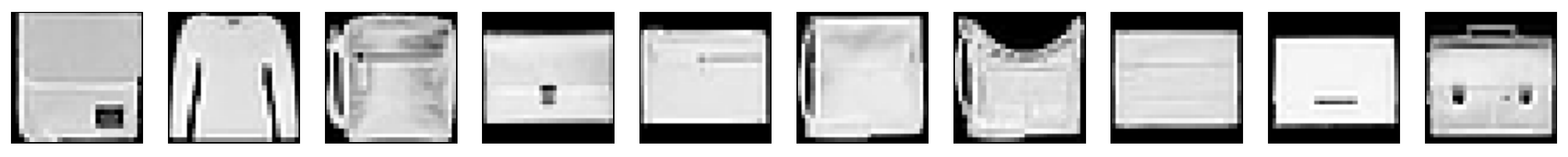} \\ 
            \small LeNet-5 & \includegraphics[width=\linewidth]{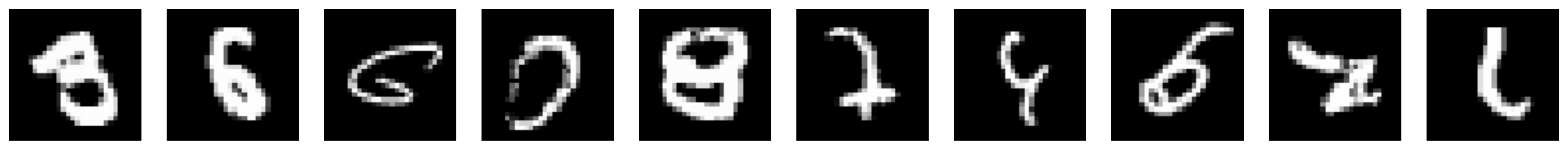}
                             \includegraphics[width=\linewidth]{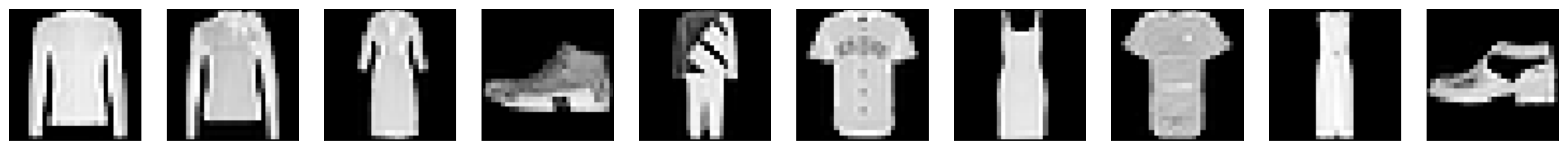}
        \end{tabular}
      \caption{Most anomalous samples in the test set for MNIST (top) and Fashion-MNIST (bottom) datasets and for each AD method and LeNet-5 classifier.}
      \label{figure-most-anomalous-samples}
    \end{center}
\end{figure}

Past works on AD with MNIST dataset arbitrarily assigned one of the classes as anomalous. For instance digit $0$ was considered abnormal while other digits were considered as inliers. We propose a different, more intuitive setup. Firstly we can subjectively asses performance of AD algorithms using test dataset simply by reporting instances regarded as most anomalous (see Figure~\ref{figure-most-anomalous-samples}). Human observer regards the digits as outliers because of the latent features not captured by class label describing the original, unconventional handwriting styles. For instance digit $4$ with style resembling digit $9$ should be considered as anomalous. To proxy this behavior we train a classifier $M$ and label each sample having classification error higher than threshold $t$ as anomalous. We apply the exact same procedure for Fashion-MNIST dataset. In our study we use LeNet-5~\cite{lecun1998gradient} model. In summary, each pre-trained AD algorithm $A$ is evaluated as in Algorithm~\ref{algoritm-ad-mnist}.

\begin{algorithm}
    \caption{AD on MNIST and Fashion-MNIST datasets}\label{algoritm-ad-mnist}
    \begin{algorithmic}[1]
        \Procedure{Label}{Model $M$, Data $X_{train}$, Data $X_{test}$}
        \State $M\gets X_{train}$ \Comment{Training Classifer}
        \State $s = M(X_{test})$ \Comment{Evaluate Log Loss}
        \State \textbf{return} $s$
        \EndProcedure
        
        \Procedure{Detection}{Algo $A$, Data $X_{test}$}
        \State $t = 0.01$
        \While{$t < 1$}
        \State $labels\gets s > t$ \Comment{Get Binary Labels}
        \State $scores\gets A(X_{test})$ \Comment{Get Anomaly Score}
        \State $p\gets AUC(labels,scores)$ \Comment{Get ROC AUC}
        \State $t = t + 0.01$
        \EndWhile
        \State \textbf{return} $p$
        \EndProcedure
    \end{algorithmic}
\end{algorithm}

In our experimental setup we assign a class label to vector $k$ while $u$ should accommodate information about other factors of variation e.g. hand used to write a digit. The problem of detecting anomalies is analogous to Type B problem. In this case we would expect $\mu(\mathbb{D}_{\text{KL}})$ to be higher in cases of mislabelling or uncommon style.

\begin{figure*}
  \begin{center}
  \includegraphics[width=0.47\linewidth]{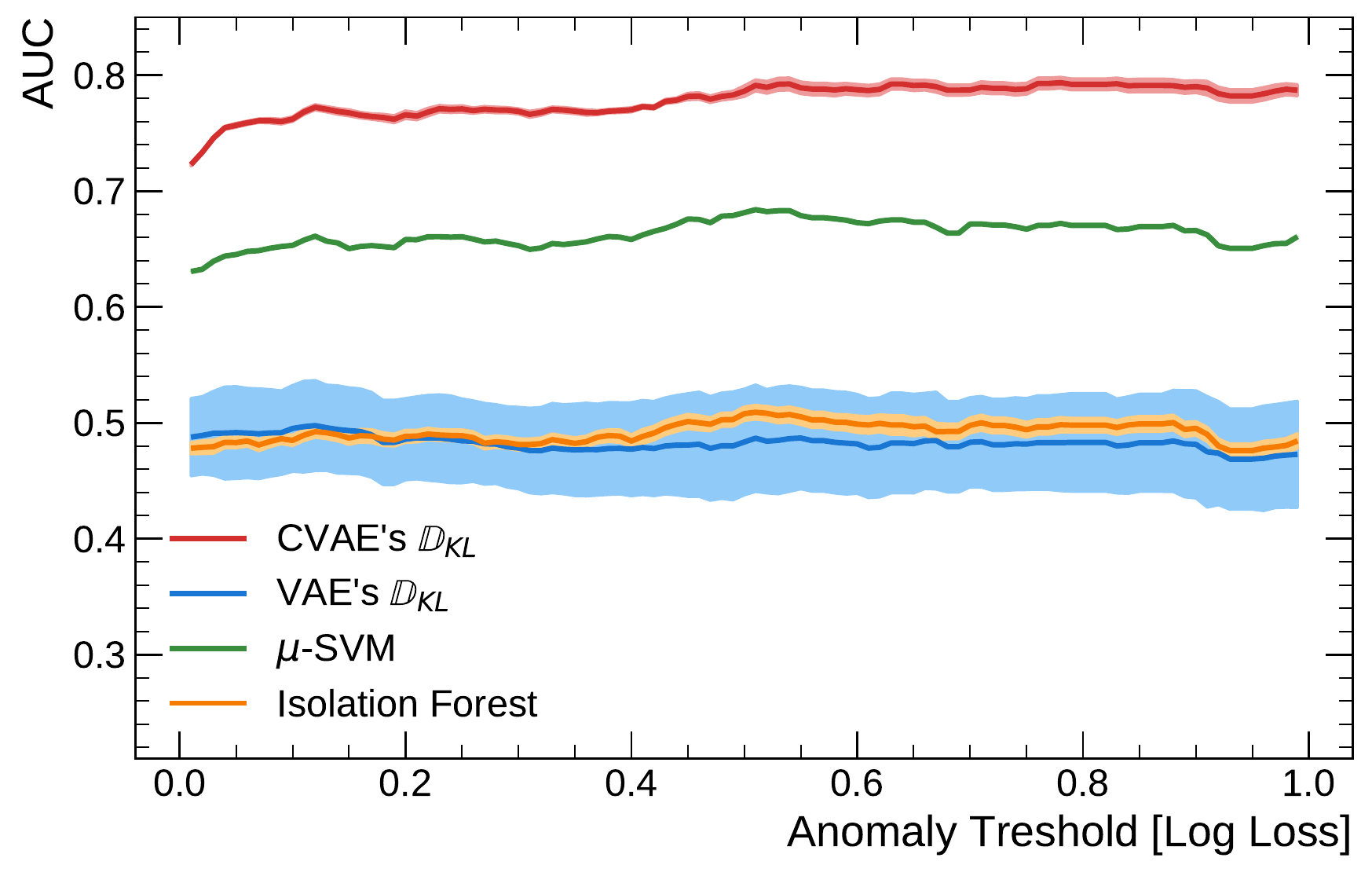}
  \includegraphics[width=0.47\linewidth]{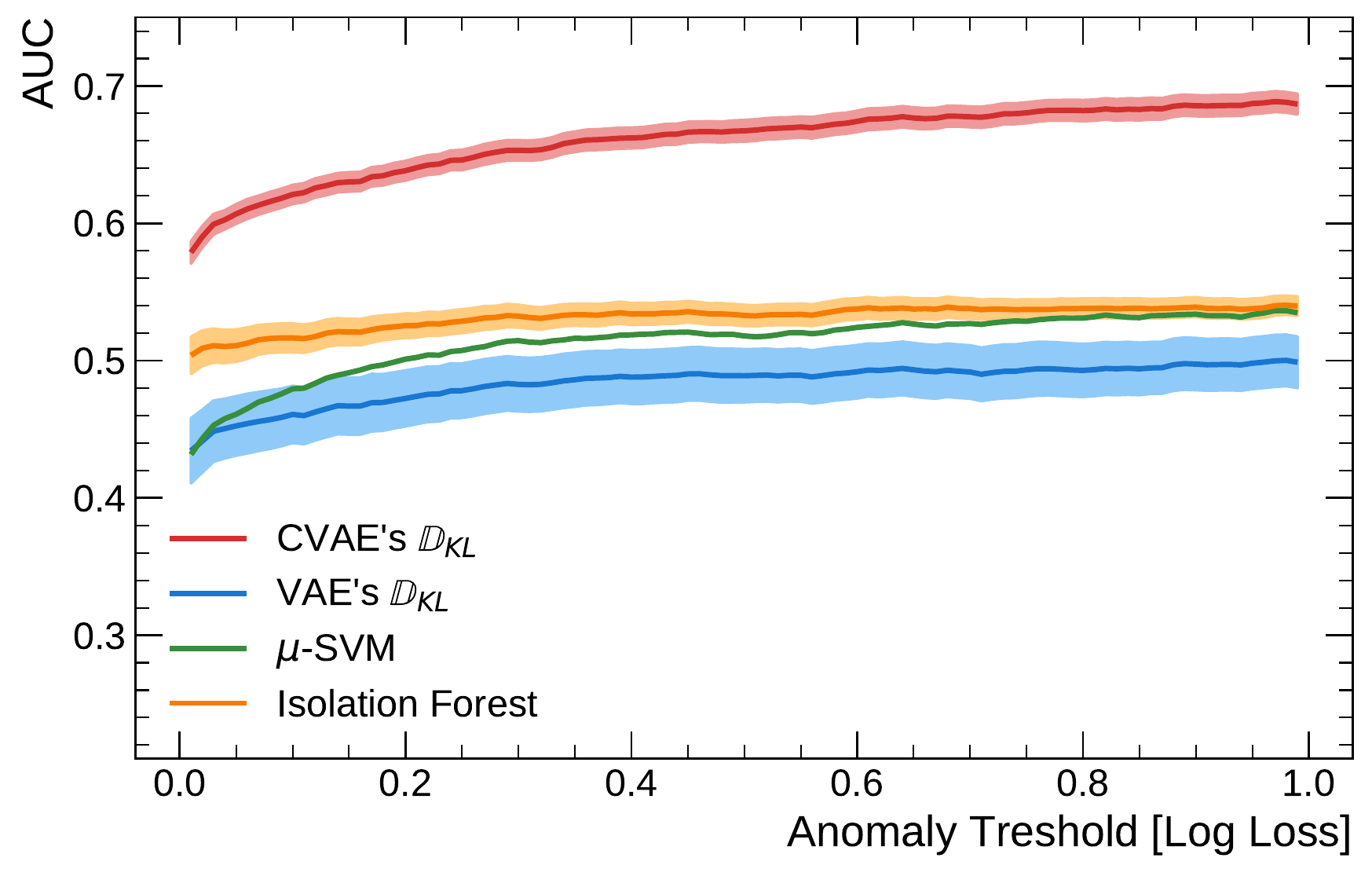}
  
  \caption{Reported ROC AUC for {\bf MNIST} (left) {\bf Fashion-MNIST} (right) datasets and different AD algorithms as a function of varying anomaly threshold $t$ based on LeNet-5 model classification log loss $s$. Overall classifier accuracy is 98.95\% and 89.62\% for MNIST and Fashion-MNIST respectively. The curves stay relatively flat due to high performance of the classifier: most of the test samples have log loss smaller than $0.01$.}
  \label{figure-mnist-roc-auc}
  \end{center}
\end{figure*}

Throughout the experiments, we use the original train-test splits with $10000$ test samples. For changing classification error threshold values $t$ we report Receiver Operating Characteristic (ROC) Area Under the Curve (AUC) $p$ for popular AD algorithms and a vanilla VAE, see Figure~\ref{figure-mnist-roc-auc}. We use $\mu$-SVM and IF as baselines, for which we concatenate class label to input pixel values for fair comparison. We notice that for vanilla VAE the $\mathbb{D}_{\text{KL}}$ is not a useful anomaly indicator, as we expect the latent information to be mostly dominated by the class-label value. Changing architecture to CVAE turns $\mathbb{D}_{\text{KL}}$ to anomaly indicator, which outperforms other baseline techniques. The Fashion-MNIST dataset was designed to be more challenging replacement for MNIST. We notice observable drop in ROC AUC as the dataset has more ambiguity between classes. However, compared to baseline methods the CVAE-based model exceeds their detection performance.

For generative purposes our setup is insufficient. As shown in~\cite{mathieu2016disentangling} we would need additional adversarial system for such objective. However, the AD task is in fact simpler as it is not necessary to generate realistic outputs of the generator. Such regularization will not help with training set contamination with outliers. This can give the encoder possibility to store too much information about the anomalies and harm the detection performance of the algorithm.

\subsection{Synthetic Problem}\label{section-synthetic}

The synthetic dataset uses normally distributed ($\mu = 0$, $\sigma = 1$), continues and independent latent variables $u$ and $k$. Observable $x$ is simply a product of $u$, $k$ and additional noise $\epsilon$ given configuration constraints: $x_{j} = f_{j}(\vec{u}) \cdot \sum_{i=0}^{m} \mathbf{S}_{ji}k_{i} + \epsilon,$ where $j$ is a feature index for $\vec{x}$ in $\mathbb{R}^{n}$. A binary matrix $\mathbf{S}$ describes which $k$ is used to compute feature $j$:

\begin{equation*}
    \mathbf{S} =  \begin{blockarray}{ccccc}
    & k_{0} & k_{1} & \dotsm & k_{m}\\
    \begin{block}{c(cccc)}
      x_{0} & 1 & 0 & \dotsm & 0 \\
      x_{1} & 1 & 0 & \dotsm & 0 \\
      \vdots & \vdots & \vdots & \ddots & \vdots \\
      x_{n} & 0 & 0 & \dotsm & 1 \\
    \end{block}
    \end{blockarray},
\end{equation*}
and function $f(\vec{u})$ describes which $u$ enters the product that defines each feature $j$: $ f_{j}(\vec{u})= \prod_{o} u_{o}.$ $S$ and $f(\vec{u})$ stay unchanged across each sample in the dataset but the values of $k$ and $u$ do change. For simplicity, we ensure that each $j$ depends only on one $k$ and the dependence is equally distributed. Finally we can manipulate values of $o$ and $m$. For instance, the first column $x_{0}$ can use $k_{0}$, $u_{1}$ and $u_{4}$: $x_{0} = k_{0}u_{1}u_{4}$; $x_{99}$ may be generated using $k_{4}$ and $u_{0}$ etc.

\begin{figure}
    \begin{center}
      \includegraphics[width=0.75\linewidth]{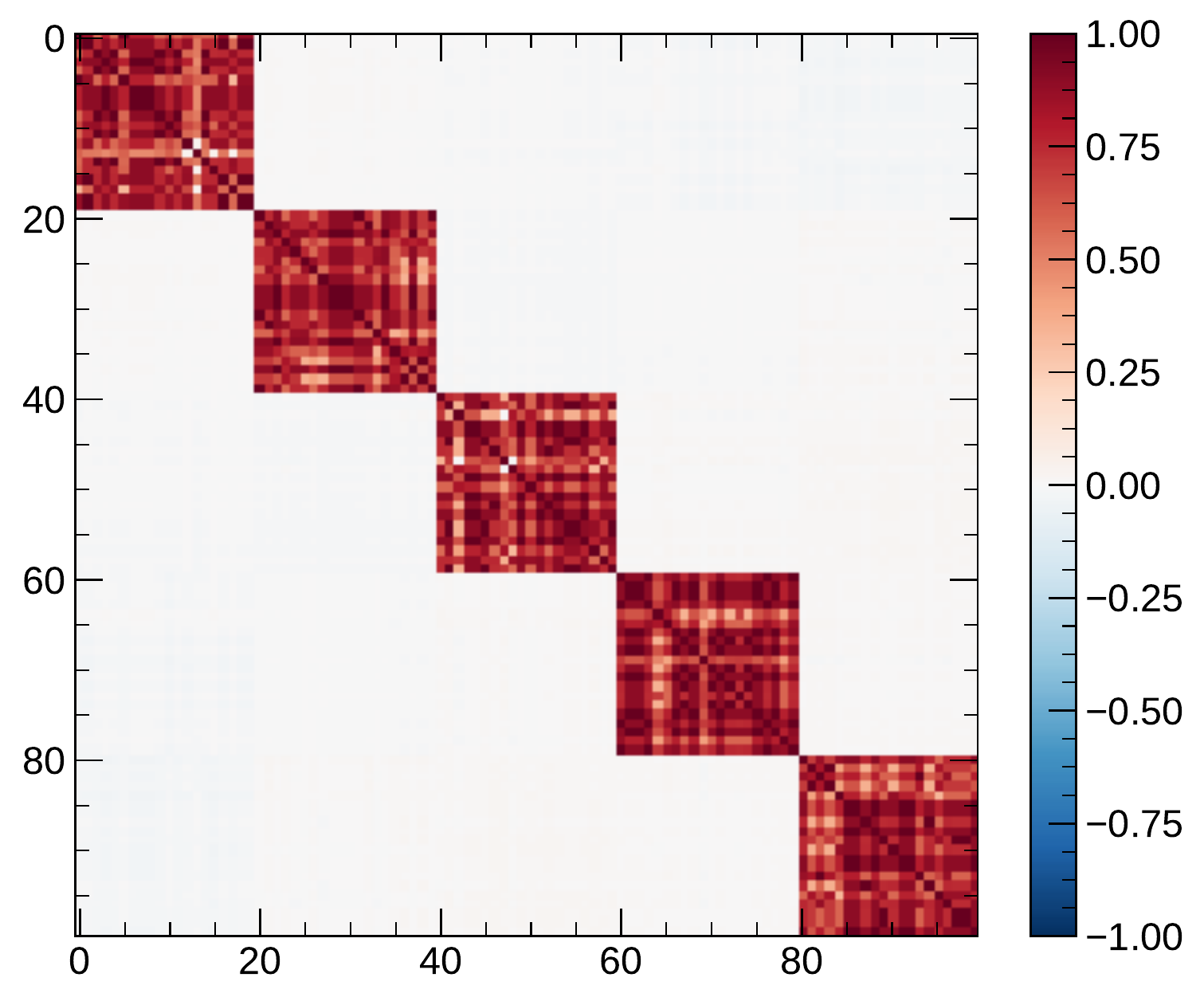}
      \caption{Correlations between features for $m=o=5$ and $n=100$.}
      \label{figure-correlation-matrix}
    \end{center}
\end{figure}

We generate samples with $x$ being $100$-dimensional ($n=100$) and $m=o=5$. An example of correlation matrix between features can be seen in Figure~\ref{figure-correlation-matrix}. For testing we generate samples according to Table~\ref{table-types-of-anomalies}. The choice of $5\sigma$ and $3\sigma$ comes from legacy requirements of our target application. The AD is performed by: comparing output of the decoder with encoder input for problems observed only on one of the features - Type A problem; or comparing $\mathbb{D}_{\text{KL}}$ yield for a samples with problems present on all features belonging to the same causal group (using the same $k$ column as input) - Type B problem.

\begin{table}
  \begin{center}
    \caption{Types of test data}
    \begin{tabular}{l|l}
      \bf{Test set} & \bf{Description} \\
      \hline
      Type A Inlier  & Generated in the same process as training data \\
      Type A Anomaly & $5\sigma$ change on $\epsilon$ for a random feature \\
      Type B Inlier  & $3\sigma$ change on $\epsilon$ for a random set of $\frac{n}{m}$ features \\
      Type B Anomaly & $3\sigma$ change on $\epsilon$ for a random feature cluster \\
    \end{tabular}
    \label{table-types-of-anomalies}
  \end{center}
\end{table}

\begin{figure*}
    \begin{center}
      \includegraphics[width=.47\linewidth]{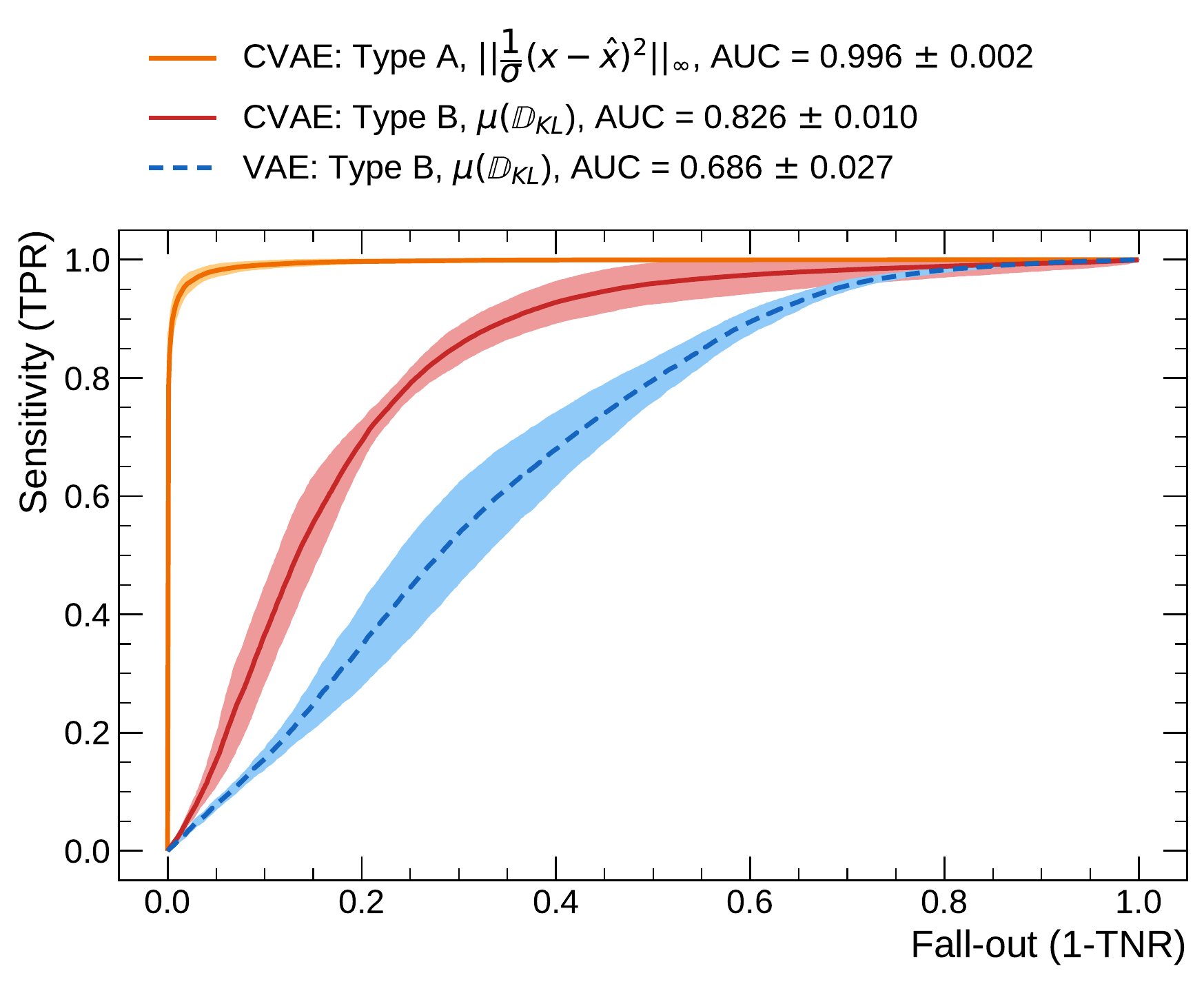}
      \includegraphics[width=.47\linewidth]{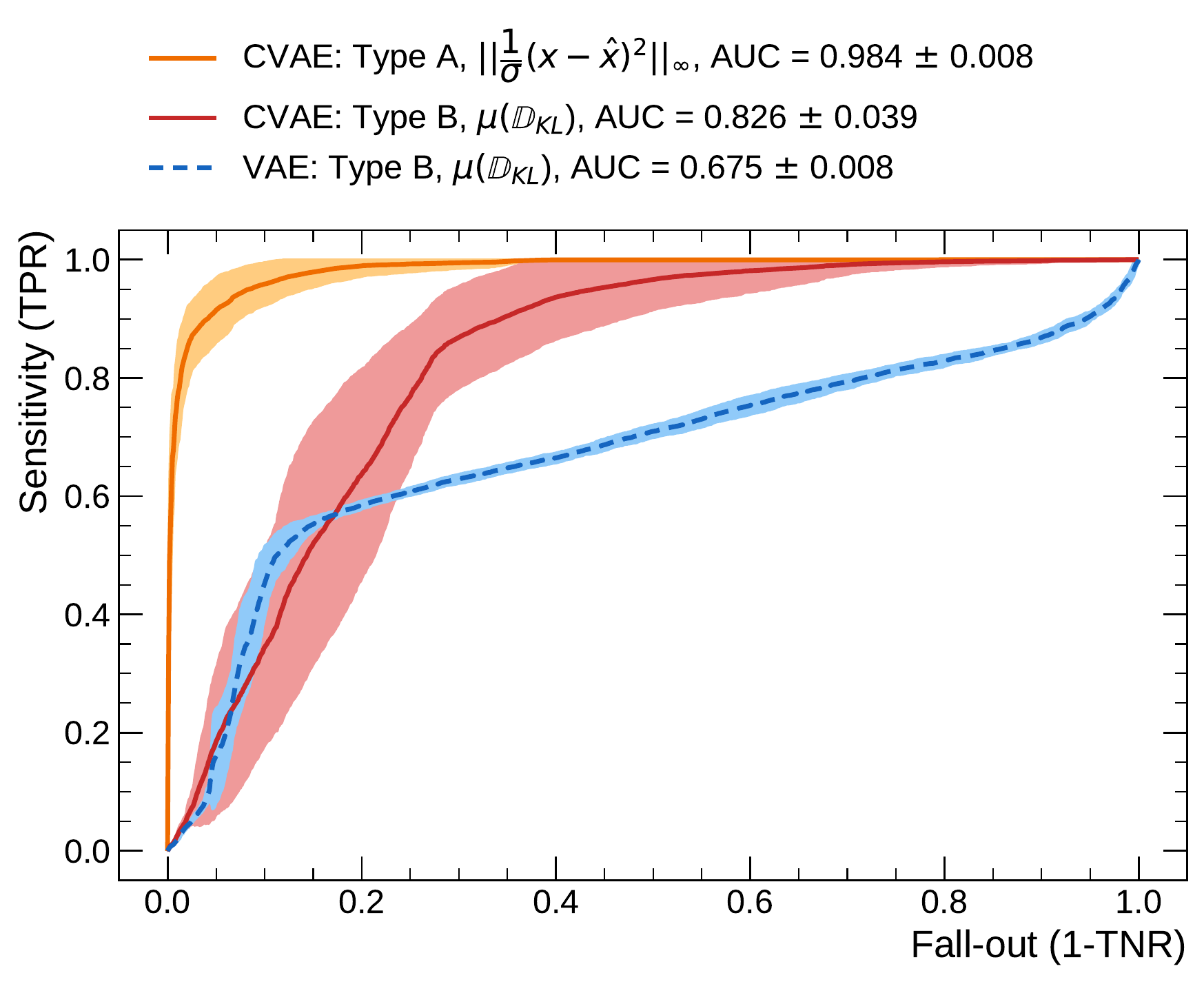}

      \caption{The ROC curves for two AD problems using synthetic test dataset (left) and CMS trigger rates test dataset (right). The bands correspond to variance computed after running the experiment five times using random weight initialization. Anomaly score for Type B is computed using mean $\mathbb{D}_{\text{KL}}$ of $z$. Anomaly score for Type A problem is computed using decoder outputs: $\mu$ and $\sigma$ of each feature. For CMS trigger case with low fall-out, VAE slightly outperforms CVAE which could be caused by our specific choice of HLT paths.}
      \label{figure-synthetic-cms-roc}
    \end{center}
\end{figure*}

The ROC curves corresponding to both of the problems are shown in Figure~\ref{figure-synthetic-cms-roc}. Given the high order of the deviation on Type A anomalies, the algorithm easily spots those types of problems. In context of hierarchical structures, an algorithm needs to model a mapping from single input to multiple possible outputs. As argued in~\cite{sohn2015learning} we need a model that can make diverse predictions. The Type B detection provides good results outperforming vanilla VAE baseline confirming that CVAEs are suitable for such task.

\section{Experiments on CMS Trigger Rate Monitoring}\label{section-experiment-cms}

\subsection{Motivation}

This work emerges directly from the explicit urgency of extending monitoring of the CMS~\cite{chatrchyan2008cms} experiment. The CMS experiment at CERN LHC~\cite{lhc1995large} operates at the remarkable rate of $40$ million particle collisions ({\em events}) per second. Each event corresponds to around $1$~MB of data in unprocessed form. Due to understandable storage constrains and technological limitations (e.g. fast enough read-out electronics), the experiment is required to reduce the number of recorded data from $40$ million to $1000$ events per second in real time. To this purpose, a hierarchical set of algorithms collectively referred to as the {\em trigger system} are used to process and filter the incoming data stream which is the start of the physics event selection process.

Trigger algorithms \cite{Khachatryan:2016bia} are designed to reduce the event rate while preserving the physics reach of the experiment. The CMS trigger system is structured in two stages using increasingly complex information and more refined algorithms:
\begin{enumerate}
    \item \textbf{The Level 1} (L1) \textbf{Trigger}, implemented on custom designed electronics; reduces the $40$~MHz input to a $100$~kHz rate in $<10~\mu$s.
    \item \textbf{High Level Trigger} (HLT), a collision reconstruction software running on a computer farm; scales the $100$~kHz rate output of L1 Trigger down to $1$~kHz in $<300~$ms.
\end{enumerate}
Both the L1 and the HLT systems implement a set of rules to perform the selection (called {\em paths}). The HLT ones are seeded by the events selected by a configurable set of L1 Trigger paths.

Under typical running conditions, the trigger system regulates the huge data deluge coming from the observed collisions. The quality of the recorded data is guaranteed, by monitoring each detector subsystems independently (e.g. measuring voltage), and by monitoring the trigger rates. The event acceptance rate is affected in presence of number of issues e.g. detector malfunctions, software problems etc. Depending on the nature of the problem, the rate associated to specific paths could change to unacceptable levels. Critical cases include dropping to zero or increasing to extreme values. In those cases, the system should alert the shift crew, calling for a problem diagnosis and intervention.

HLT paths are often very strongly correlated. This is due to the fact that groups of paths select similar physics objects (thus reconstructing the same event) and/or are seeded by the same selection of L1 Trigger paths. While critical levels of rate deviations for singular paths should be treated as anomaly, smaller deviations on number of random trigger paths are likely a result of statistical fluctuations. On the other hand an observable coherent drift (even small) on a group of trigger paths related by similar physics or making use of the same hardware infrastructure, is an indication of a likely fault present in the trigger system or hardware components. 

\begin{figure*}
    \begin{center}
      \includegraphics[width=0.7\linewidth]{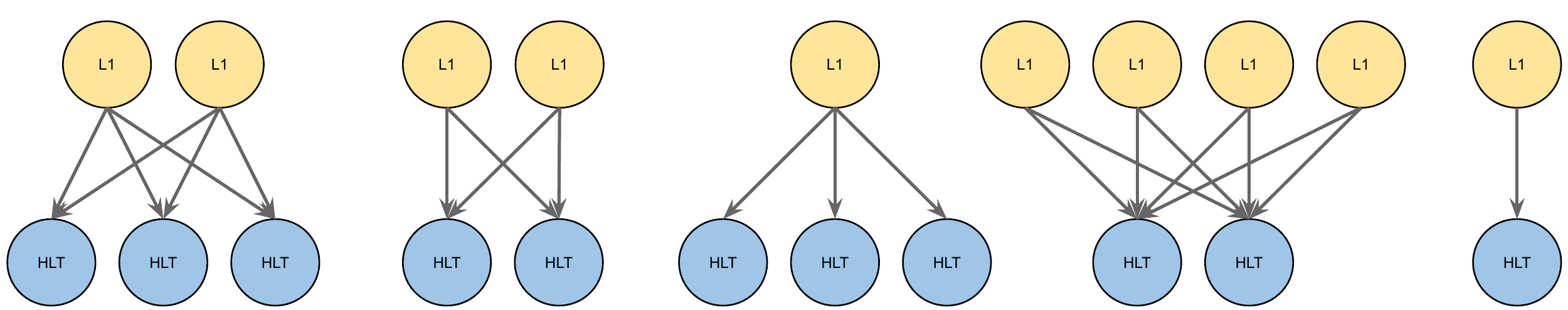}
      \caption{Simplified, schematic graph inspired by the trigger system configuration. Blue nodes represent HLT paths while yellow L1 trigger paths. Each link is unidirectional starting from yellow nodes. The graph has few hundred nodes spread approximately equally between HLT and L1 triggers paths. The connection between L1 trigger and HLT paths can be seen as a hierarchical directional graph from L1 to HLT system.}
      \label{figure-schematic-configuration}
    \end{center}
\end{figure*}

We explore this hierarchical structure in our algorithm. Each HLT path has a direct, pre-configured link to set of L1 trigger paths through specified configuration as schematically shown in Figure~\ref{figure-schematic-configuration}. The configuration changes infrequently i.e. nodes are added, disabled or corrected. Consequently, the HLT system performance is directly linked with the status of L1 Trigger.

We do not focus on minimizing the inference time as the anomaly can be flagged within minutes which is long enough for all the algorithms considered.

\subsection{Experiment}

We apply CVAE architecture, where we treat HLT rates as $x$ and L1 Trigger rates as $k$. Our prototype uses four L1 Trigger paths that seed six unique HLT paths each. We extract rates only from samples where all chosen paths are present in the configuration. We end up with $102895$ samples which are then split into training, validation and test set. Our test set has 2800 samples. Operators set quality labels for each CMS sub-detector and for each sample. Since the global quality flag is composed by contribution from all subsystems, a sample could be regarded as bad due to under-performance of a detector component not related to the set of trigger paths we chose or not related to problem we try to solve. Hence we cannot use those labels in the test set. Instead, we consider hypothetical situations that are likely to happen in the production environment, similar to those used for synthetic problem. We generate four synthetic test datasets manipulating our test set in similar manner to the synthetic dataset. We detect isolated problems on one of the HLT paths - Type A; and problems present across HLT paths seeding the same L1 trigger path - Type B.

We report the results in Figure~\ref{figure-synthetic-cms-roc}. The performance of the algorithm on CMS dataset is matching the performance we reported for the synthetic one. The CMS experiment currently does not provide any tools to track problems falling into Type B category. Given a good performance of the proposed method, we believe that the solution could be considered for deployment, provided further tests and refinements in the production environment.

\section{Conclusions and Future Work}

This paper shows how anomalous samples can be identified using CVAE. We considered the specific case of CMS trigger rate monitoring to extend the current monitoring functionality and showed good detection performance. The proposed algorithm does not rely on synthetic anomalies at training time or additional feature engineering. We demonstrated the method is not bound to CMS experiment specifics and has potential to work across different domains. However more tests on more difficult datasets are desirable, e.g. on CIFAR, which provides more classes and a higher variance. We did not perform hyper-parameter scan for any of the experiments thus we expect the results to get better if further optimized. Subsequent studies foresee using full configuration of the CMS trigger system. An interesting extension of the method would be learning correct encoding of unknown factors of variations in the latent space, which at this moment is unconstrained (e.g. a tilt or boldness of the digit in the MNIST dataset).

\subsubsection*{Acknowledgments}
We thank the CMS collaboration for providing the dataset used in this study. We are thankful to the members of the CMS Physics Performance and Dataset project for useful discussions and suggestions. We acknowledge the support of the CMS CERN group for providing the computing resources to train our models. This project has received funding from the European Research Council (ERC) under the European Union's Horizon 2020 research and innovation program (grant agreement n$^o$ 772369).

\bibliographystyle{unsrt}
\bibliography{paper}

\end{document}